\documentclass{article}

\usepackage[square, numbers]{natbib}
\usepackage[final]{neurips_2023}
\usepackage{wrapfig}
\usepackage{caption}
\usepackage{capt-of}

\usepackage[utf8]{inputenc} 
\usepackage[T1]{fontenc}    
\usepackage{hyperref}       
\usepackage{url}            
\usepackage{booktabs}       
\usepackage{amsfonts}       
\usepackage{nicefrac}       
\usepackage{microtype}      
\usepackage{xcolor}         

\usepackage{graphicx}
\usepackage{microtype}
\usepackage{multirow}
\usepackage{amsmath}

\graphicspath{{./figs/}}

\title{Solving Math Word Problems with Reexamination}

\author{
  Yi Bin\textsuperscript{1}, Wenhao Shi\textsuperscript{2}, Yujuan Ding\textsuperscript{3}, Yang Yang\textsuperscript{2}, See-Kiong Ng\textsuperscript{1}\\
  \textsuperscript{1}National University of Singapore\\
  \textsuperscript{2}University of Electronic Science and Technology of China \\
\textsuperscript{3}The Hong Kong Polytechnic University \\
}

\newcommand{\myeg}{\emph{e.g.}}
\newcommand{\myie}{\emph{i.e.}}
\newcommand{\myec}{\emph{etc}}

\begin{document}

\maketitle

\begin{abstract}

Math word problem (MWP) solving aims to understand the descriptive math problem and calculate the result, for which previous efforts are mostly devoted to upgrade different technical modules. This paper brings a different and novel perspective of \textit{reexamination process} during training by introducing a pseudo-dual task to enhance the MWP solving. We propose a pseudo-dual (PseDual) learning scheme to model such process, which is model-agnostic thus can be adapted to any existing MWP solvers.  The pseudo-dual task is specifically defined as filling the numbers in the expression back into the original word problem with numbers masked. To facilitate the effective joint learning of the two tasks, we further design a scheduled fusion strategy for the number infilling task, which smoothly switches the input from the ground-truth math expressions to the predicted ones. Our pseudo-dual learning scheme has been tested and proven effective when being equipped in several representative MWP solvers through empirical studies. \textit{The codes and trained models are available at:} \url{https://github.com/steven640pixel/PsedualMWP}. 
\end{abstract}

\section{Introduction}
\label{sec:intro}

Math Word Problem (MWP) solving is  to understand descriptive mathematical problems and reason the results with proper  arithmetic expression. 
Despite the great progress has been achieved with the deep learning, pre-trained language models (PLMs), and recent large language models(LLMs), the MWP solving is still challenging and under-explored. In this paper, we propose a new perspective to enhance it by introducing the \textit{reexamination process of humans} in a pseudo-dual learning scheme. The idea is inspired by the natural process of human beings addressing the MWPs, in which a reexamination procedure is appreciated to verify the correctness of the MWP solutions. 
Such solving and reexamining process formulates a closed loop and has also been in many areas, e.g., dual-learning for machine translation~\cite{he2016dual} and dual/cycle-learning methods for image2image translation~\cite{yi2017dualgan,zhu2017unpaired,zheng2022asynchronous}, further showing the potential of our proposed idea to enhance the MWP solving problem. 

To this end, we try to design the dual problem and a proper joint learning scheme which could positively influence the main MWP solving task. The strict dual problem, \myie, the reverse process, of MWP solving is generating the original problem description based on the mathematical expression. However, the success of solving the MWP relies mostly on the capture of numbers in the problem description and the mathematical logic among them rather than the detailed comprehension of descriptive text. From another perspective, reconstructing the problem description simply based on the mathematical expression is unbelievably challenging, even more challenging than the main task, which could bring negative effect instead. For these considerations, in this paper, we introduce a relaxed reverse process, dubbed pseudo-dual learning scheme (PseDual), filling the numbers into the masked problem description based on given mathematical expressions, which can be regarded as the pseudo-dual task for the main MWP solving task. The PseDual scheme is model-agnostic and has been applied to several representative MWP solving models in this work, \myeg, DNS~\cite{wang2017deep}, Graph2Tree~\cite{zhang2020graph}, BERT-Tree~\cite{Li2022aclcl}, \myec, and proven its effectiveness. For the effective joint learning of the main and pseudo-dual tasks, we further propose a scheduled fusion learning strategy, which is motivated by curriculum learning~\citep{bengio2009curriculum}. Specifically, in the process of the joint training, the number infilling part applies the ground-truth mathematical expression as input in the beginning and gradually switches to the predicted expression. On one hand, the predicted expression is hardly ideal at the beginning of the training stage, which would mislead the pseudo-dual task learning and thus is not proper to be applied as input. On the other hand, such a scheduled fusion strategy can gently balance the two learning parts during training by adaptively adjusting the `weight' between them with different scales of losses. The main contributions of this paper can be summarized as follows:

\begin{itemize}
\item This paper proposes to investigate the \textit{reexamination process} for MWP solving and introduces a novel Pseudo-Dual (PseDual) Learning scheme, which provides a new perspective on this topic and is model-agnostic. It implements a relaxed dual task, \myie{}, number infilling, to jointly learn with the main MWP solving task and further enhance its accuracy.

\item To jointly train the solving module and reexamining module, we devise a scheduled fusion strategy to smoothly switch the infilling expression from ground-truths to predictions.

\item Extensive experiments are conducted on three datasets based on several representative models, and the results demonstrate the effectiveness of the proposed approach. Besides, we also investigate the integration of reexamination process with LLMs, \myeg{}, ChatGPT, further verifying its effectiveness and generalization ability.
\end{itemize}

\section{The Proposed Approach}
\label{sec:app}

\subsection{Model Architecture}

As aforementioned, the proposed PseDual framework for MWP solving mainly consists of two cycled modules: \textbf{solving module} for solution expression generation, and \textbf{reexamining module} for verifying the correctness of expression by filling the numbers into the masked problem.

\textbf{Solving Module.} Following previous works, the solving module contains the word problem encoder and expression generation decoder, each could be implemented by different models. In this work we employ several representative encoder and decoder models, and establish different solving modules by combining different encoders and decoders. Specifically, for encoder we have RNN-based and pretrained language models (PLMs), and  sequential and binary tree models for decoder.

\textbf{Reexamining Module.} After the solving module generates the expression $S=\{s_1, s_2, ..., s_k\}$, humans always attempt to reexamine the correctness of it. Inspired by the success of dual-/cycle- learning mechanism in machine translation and image translation~\cite{he2016dual,yi2017dualgan,zhu2017unpaired}, a straightforward way for reexamination is to ``translate'' the expression to the original problem to enhance the ability of understanding and reasoning of solving module. However, an expression could be associated with multiple problems,  which makes the problem reconstruction from expression extremely challenging. From another perspective, the quantities in the expression can be matched to the numbers at the problem, and the operators denote the relations described in the problem~\cite{jie2022learning}. Besides, the information beyond the numbers and relations contributes very little to the expression. To this end, 
we relax the task from \underline{\textit{reconstructing problems based on the expressions}} to \underline{\textit{filling the numbers in the problems based on the expressions}}. Such a pseudo-dual task design emphasizes the capturing of mathematical relations in the expressions and problems. In fact, training the model for filling masked blanks in sentence/paragraph is an effective way to enhance the modeling of context and understanding of the crucial information in NLP~\cite{BERT,bin2021entity,donahue2020enabling}. Motivated by the expression representation in the MWP solving works, we employ two kinds of architectures for expression encoder: sequential model and binary tree.

\subsection{Scheduled Fusion}
\label{sec:scheduled}
We note that it is hard to jointly train the solving module and reexamining module from scratch, because the predicted expression in the beginning is far from ideal for infilling, which would mislead the training process of the whole model. One feasible solution is teacher forcing~\cite{Bengio_schedule_nips}, which utilizes the ground-truth expression as input of reexamination, but exists a gap between training and test. Motivated by~\cite{bengio2009curriculum}, we propose a novel scheduled fusion strategy to address this issue. Specifically, during training, we adopt the integration of ground-truth and predicted expressions to obtain infilling number representations, and introduce a weight $\epsilon$ to balance the  proportion of each component as:
\begin{equation}
    Q = \epsilon Q_g+(1-\epsilon)Q_p,
    \label{eq:sf}
\end{equation}
where the $Q_g$ and $Q_p$ are representations of quantities derived from ground-truth and predicted expressions, and $\epsilon \in [0,1]$ is adaptively adjusted by an exponential decay following~\cite{Bengio_schedule_nips,Liu_schedule_emnlp}. With such scheduled fusion, the reexamining module takes more information from ground-truth expression as input in the beginning and smoothly ``switches'' to the predicted expression.

\section{Experiments and Result Analyses}
\label{sec:exp}

\textbf{Expression and Value Accuracy.}
To verify the effectiveness of the proposed reexamination process, we conduct experiments on three commonly used datasets: Math23k \cite{wang2017deep}, MathQA \cite{Amini2019mathqa}, and MAWPS~\cite{koncel2016mawps}.
As the results shown in Table~\ref{tab:baseline}, we observe that introducing \textit{reexamination} with our proposed PseDual learning scheme, both expression and value accuracy of all the MWP solvers are significantly improved on all the datasets, \myeg{}, boosting 1.14, 0.98, and 0.88 on average
for the answer accuracy on Math23k, MathQA, and MAWPS.
Such improvements absolutely verify the effectiveness and robustness of the proposed pseudo-dual learning scheme for MWP solving. We also note that encoding expressions with GCN performs much better than GRU, because the GCN takes the expression tree as input while the GRU regards an expression as token sequence and ignores the arithmetical architecture in it. 
This observation is also consistent with that binary tree structure is more suitable and reasonable for expression representation than sequence.

\begin{table*}
\small

\centering
\caption{Experimental results with/without our PseDual scheme. ${\dagger}$ and $\ast$ are reported results and our reproduced results with the released code, respectively.
The GRU and GCN in brackets denote the encoders for expressions.}

\setlength{\tabcolsep}{0.9mm}{
\begin{tabular}{c|cc|cc|cc}
\hline
\multicolumn{1}{c|}{{\multirow{2}{*}{\textbf{Model}}}} & \multicolumn{2}{c|}{\textbf{Math23k}} & \multicolumn{2}{c|}{\textbf{MathQA}} & \multicolumn{2}{c}{\textbf{MAWPS}}   \\ 
\multicolumn{1}{c|}{} & \textbf{Expression}  & \textbf{Value} & \textbf{Expression} & \textbf{Value} & \textbf{Expression} & \textbf{Value}         \\ \hline
    DNS$^{\dagger}$          & - & 58.1 & - & - & - & 59.5   \\
    DNS$^{\ast}$          & 52.1 & 58.6 & 65.4 & 65.7 & 59.2 & 59.6   \\
    DNS+PseDual (GRU)          & 52.8 & 59.9 & 66.2 & 66.7 & 60.2 & 60.7   \\
    DNS+PseDual (GCN)          & 53.1 \footnotesize{\textbf{(${\uparrow}$1.0)}} & 60.2 \footnotesize{\textbf{(${\uparrow}$1.6)}} & 66.5 \footnotesize{\textbf{(${\uparrow}$1.1)}} & 67.1 \footnotesize{\textbf{(${\uparrow}$1.4)}} & 60.4 \footnotesize{\textbf{(${\uparrow}$1.2)}} & 61.0 \footnotesize{\textbf{(${\uparrow}$1.4)}}  \\ \hline
    GTS$^{\dagger}$          & - & 75.6 & - & 71.3 & - & 82.6   \\
	GTS$^{\ast}$          & 64.2 & 75.6 & 68.6 & 71.2 & 81.9 & 82.7   \\
	GTS+PseDual (GRU)           & 65.2 & 76.4 & 68.9 & 71.9 & 82.5 & 83.5   \\
	GTS+PseDual (GCN)          & 65.4 \footnotesize{\textbf{(${\uparrow}$1.2})} & 76.7 \footnotesize{\textbf{(${\uparrow}$1.1)}} & 69.1 \footnotesize{\textbf{(${\uparrow}$0.5)}} & 72.1 \footnotesize{\textbf{(${\uparrow}$0.9)}} & 82.6 \footnotesize{\textbf{(${\uparrow}$0.7)}} & 83.6 \footnotesize{\textbf{(${\uparrow}$0.9)}}   \\ \hline
	Graph2Tree$^{\dagger}$          & - & 77.4 & - & 72.0 & - & 83.7   \\
	Graph2Tree$^{\ast}$           & 65.5 & 77.4 & 68.9 & 72.0 & 82.6 & 83.7   \\
	Graph2Tree+PseDual (GRU)         & 66.1 & 78.0 & 69.7 & 72.7 & 83.2 & 84.4   \\
	Graph2Tree+PseDual (GCN)          & 66.4 \footnotesize{\textbf{(${\uparrow}$0.9)}} & 78.3 \footnotesize{\textbf{(${\uparrow}$0.9)}} & 70.0 \footnotesize{\textbf{(${\uparrow}$1.1)}} & 72.9 \footnotesize{\textbf{(${\uparrow}$0.9)}} & 83.6 \footnotesize{\textbf{(${\uparrow}$1.0)}} & 84.7 \footnotesize{\textbf{(${\uparrow}$1.0)}}  \\ \hline
	BERT-Tree$^{\dagger}$          & 71.2 & 82.4 & 73.5 & 75.1 & - & -  \\
	BERT-Tree$^{\ast}$           & 71.1 & 82.3 & 73.7 & 75.5 & 88.1 & 88.7  \\
	BERT-Tree+PseDual (GRU)          & 71.7 & 83.6 & 74.5 & 76.6  & 88.7 & 89.4 \\
	BERT-Tree+PseDual (GCN)          & 71.8 \footnotesize{\textbf{(${\uparrow}$0.7)}} & 84.1 \footnotesize{\textbf{(${\uparrow}$1.8)}} & 74.7 \footnotesize{\textbf{(${\uparrow}$1.0)}} & 76.9 \footnotesize{\textbf{(${\uparrow}$1.4)}} & 88.8 \footnotesize{\textbf{(${\uparrow}$0.7)}} & 89.5 \footnotesize{\textbf{(${\uparrow}$0.8)}}  \\ \hline
	RE-Deduction$^{\dagger}$         & - & 84.3 & - & 78.6 & - & 92.0   \\
	RE-Deduction$^{\ast}$           & 77.2 & 84.3 & 72.1 & 78.6 & 88.9 & 92.1   \\
	RE-Deduction+PseDual (GRU)          & 77.5 & 84.5 & 72.3 & 78.8 & 89.0 & 92.3  \\
	RE-Deduction+PseDual (GCN)          & 77.6 \footnotesize{\textbf{(${\uparrow}$0.4)}} & 84.6 \footnotesize{\textbf{(${\uparrow}$0.3)}} & 72.4 \footnotesize{\textbf{(${\uparrow}$0.3)}} & 78.9 \footnotesize{\textbf{(${\uparrow}$0.3)}} & 89.3 \footnotesize{\textbf{(${\uparrow}$0.4)}} & 92.4 \footnotesize{\textbf{(${\uparrow}$0.3)}}  \\
	 \hline
\end{tabular}
}
\label{tab:baseline}
\end{table*}

\begin{wraptable}{r}{5cm}
\small
\centering
\caption{Results of the integration of reexamination process and LLMs.}
\begin{tabular}{l|c}
\hline
     \textbf{Model} & \textbf{Value Acc}          \\ \hline
    RE-Deduction$^{\dagger}$~\cite{jie2022learning} & 45.0 \\
    Zero-Shot CoT~\cite{kojima2022large}           & 63.7  \\ \hline
    ChatGPT & 69.3\\
    ChatGPT+PseDual           & 71.8  \\ \hline

\end{tabular}
\label{tab:llm}
\end{wraptable}
\noindent\textbf{Investigating the reexamination with LLMs.} Recently, Large Language Models (LLMs) have led to notable advancements in many NLP problems, such as dialog and MWPs solving~\cite{wei2022chain,kojima2022large}. 
To investigate the integration of  our proposed \textit{reexamination process} with LLMs, we conduct several empirical experiments with ChatGPT
API (\texttt{gpt-3.5-turbo} in specific), on SVAMP dataset because most LLMs were tested on it. We implement a zero-shot fashion prompt to solve the MWPs and reexamine the predicted solutions. 
We show the results in Table~\ref{tab:llm}, and observe that ChatGPT exhibits superior performance than zero-shot CoT in math problem reasoning, which is consistent with the fact that ChatGPT is more powerful than previous LLMs, \myeg{}, GPT3 and PaLM. By introducing our reexamination process, the performance gains significant improvement (from 69.3 to 71.8), further verifying the effectiveness and generalization ability of the proposed reexamination process for MWP solving.

\begin{wraptable}{r}{6.cm}
\small
\centering
\caption{Comparison between scheduled fusion and teacher forcing.}
\setlength{\tabcolsep}{0.8mm}{
\begin{tabular}{l|cc}
\hline
    & \textbf{Expression} & \textbf{Value}          \\ \hline
    \multicolumn{1}{l|}{ \textbf{DNS}}  & & \\
    ~~~~Teacher Forcing   & 52.8 & 59.9  \\
    ~~~~Scheduled Fusion           & 53.1 & 60.2  \\ \hline
    \multicolumn{1}{l|}{\textbf{GTS}}  & & \\
    ~~~~Teacher Forcing           & 65.3 & 76.3  \\
    ~~~~Scheduled Fusion           & 65.4 & 76.7 \\ \hline
    \multicolumn{1}{l|}{\textbf{Graph2Tree}}  & & \\
    ~~~~Teacher Forcing           & 66.2 & 78.0  \\
    ~~~~Scheduled Fusion           & 66.4 & 78.3  \\ \hline
    \multicolumn{1}{l|}{\textbf{BERT-Tree}}  & & \\
    ~~~~Teacher Forcing           & 71.6 & 83.7  \\
    ~~~~Scheduled Fusion           & 71.8 & 84.1  \\ \hline
    \multicolumn{1}{l|}{\textbf{RE-Deduction}}  & & \\
    ~~~~Teacher Forcing           & 77.5 & 84.4  \\
    ~~~~Scheduled Fusion           & 77.6 & 84.6  \\ \hline

\end{tabular}}
\label{tab:schsamp}
\end{wraptable}
\noindent\textbf{Scheduled Fusion \textit{v.s.} Teacher Forcing.} As aforementioned, to eliminate the discrepancy between training and inference, we propose a scheduled fusion strategy for the infilling expression balancing during training. To evaluate the effectiveness of this strategy, we also implement teacher forcing strategy (with GCN as expression encoder) and conduct experiments on Math23K. Table~\ref{tab:schsamp} shows the comparison results, from which we can observe that scheduled fusion consistently outperforms the teacher forcing training. This is benefiting from that the feedback signal of teacher forcing only optimize the problem encoder in solving module, while the scheduled fusion jointly optimize all the parts in solving module, including problem encoder and expression generator, which enables the solving module to boost the ability of  problem understanding and solution reasoning. Besides, we note that the value accuracy gains more improvements than expression. This observation may imply that through our scheduled fusion, the models could correctly perceive and reason the arithmetical relations and derive the gold answer for the problem, while the predicted expression is different from the ground-truth and  evaluated to be wrong, due to there may exist multiple correct expressions for given problem.

\begin{wrapfigure}{r}{7cm}
\includegraphics[width= \linewidth]{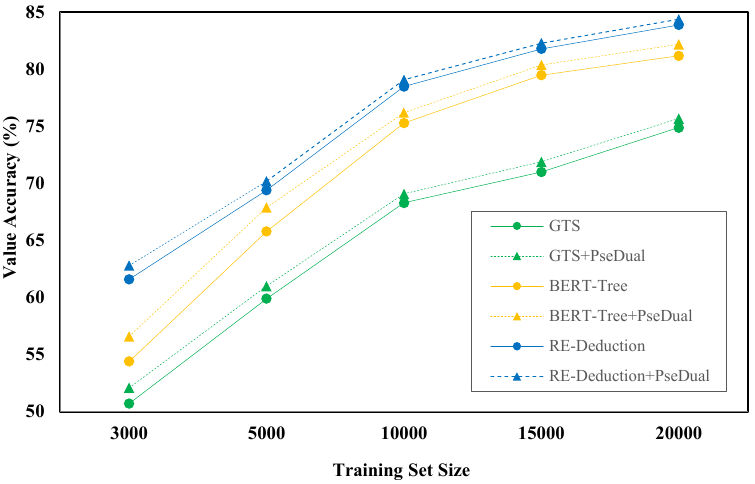}
	\caption{The performance with different training set sizes on Math23K.}
 \label{fig:tr_sz}
\end{wrapfigure}
\noindent\textbf{Different Training Set Sizes.} It is well known  that deep neural models are data-hungry and computation costly during training. To investigate what the effect of our reexamination process with PseDual to the models for different scales of training set, we conduct experiments for GTS, BERT-Tree, and RE-Deduction with training set of $\{ 3000, 5000, 10000, 15000, 20000 \}$ random samples. The validation and test sets are the same with previous experiments, and all the settings are the same as Section~\ref{sec:setting}. As the results shown in Figure~\ref{fig:tr_sz}, with the growing of the training set size, the performances with and without PseDual are increasing consistently for all the approaches, and the increase trend gets to  flat with large training set. More importantly, the improvements introduced by our pseudo-dual learning are more significant for smaller training set, which means that equipped with our PseDual would encourage the model to comprehensively understand and explore the data. In summary, our reexamination process with pseudo-dual learning demonstrates superiority at the less training samples situation.

\vspace{-5pt}
\section{Conclusion}
\vspace{-5pt}
\label{sec:con}
In this paper, we proposed a novel perspective, \textit{reexamination process}, for MWP solving, and implemented it with a pseudo-dual (PseDual) learning scheme. 
Beyond employing advanced techniques to design fancy models, we introduced the reexamination process as a pseudo-dual task to jointly learn with the MWP solving task and improve the accuracy.
The proposed PseDual scheme is model-agnostic and could be adopted to most existing MWP solving methods to further improve their performance.
Extensive experiments were conducted on three datasets, and the results demonstrated the effectiveness of the proposed PseDual scheme.

\section{Acknowledgement}
This research is supported by A*STAR, CISCO Systems (USA) Pte. Ltd and National University of Singapore under its Cisco-NUS Accelerated Digital Economy Corporate Laboratory (Award I21001E0002). This research is also partially supported by the National Natural Science Foundation of China under grant 62102070, 62220106008, and U20B2063. This research is also partially supported by Sichuan Science and Technology Program under grant 2023NSFSC1392.

\bibliography{Pseudo-Dual}

\appendix

\section{Infilling Accuracy} Although the reexamining module would not be involved in the inference, we still attempt to investigate the infilling accuracy with ground-truth expressions, which may provide  some insights to further improve the pseudo-dual learning scheme. Table~\ref{tab:infilling acc} illustrates the results, where \textbf{Acc}  means absolute accuracy over all the masked numbers in the dataset, and Perfect Match Ratio (PMR) is perfect matching ratio indicating true when all the slots in a masked problem are correctly infilled. We can observe that the results exhibit similar trend with MWP solving accuracy over different baselines, which means the basic modules, \myeg{}, BERT or BiGRU for problem encoding, are crucial for both solving and infilling, and could increase the performance together. For the best model RE-Deduction, more than 85 percent slots could be correctly infilled for all the datasets, and about 70 percent problems could be completely filled with right numbers, indicating that there remains a large space to further improve the   \textit{reexamination process}.

\begin{table*}[h]
\centering
\caption{Number infilling Accuracy and Perfect Match Ratio (PMR) of target expression.}

\setlength{\tabcolsep}{3.0mm}{
\begin{tabular}{c|cc|cc|cc}
\hline
\multicolumn{1}{c|}{{\multirow{2}{*}{\textbf{Model}}}} & \multicolumn{2}{c|}{\textbf{Math23k}} & \multicolumn{2}{c|}{\textbf{MathQA}} & \multicolumn{2}{c}{\textbf{MAWPS}}   \\ 
\multicolumn{1}{c|}{} & Acc  & PMR & Acc & PMR  & Acc & PMR        \\ \hline
    DNS          & 69.65 & 46.3 & 71.05 & 42.91 & 73.55 & 52.25   \\ 
    GTS          & 76.25 & 59.7 & 75.54 & 54.78 & 80.31 & 62.05   \\ 
    Graph2Tree   & 78.68 & 61.3 & 76.91 & 57.43 & 83.29 & 65.67   \\ 
    BERT-Tree    & 82.75 & 67.1 & 80.66 & 64.12 & 85.32 & 69.48   \\ 
    RE-Deduction & 87.82 & 72.6 & 85.45 & 69.20 & 89.11 & 73.28  \\ \hline

\end{tabular}
}
\label{tab:infilling acc}
\end{table*}

\section{Datasets and Evaluation Metrics}
We have conducted experiments on three commonly used datasets:
Math23k \cite{wang2017deep}, MathQA \cite{Amini2019mathqa} and MAWPS \cite{Koncel2016NAACL}. Math23k contains 23k Chinese math problems for primary school, including annotations of problem text, solution expression, value of answer, and number list of the problem. MathQA is an English dataset and involves more operations and complex domains, thus is more difficult to solve than Math23k. We process MathQA following \cite{Tan2021pre, Li2022aclcl, jie2022learning}  to get consistent annotation with Math23k and remove unsolvable problems in original dataset. Since \citet{Tan2021pre, Li2022aclcl} and \citet{jie2022learning} obtain different data split for MathQA, we use corresponding processed MathQA dataset under different backbone for fair comparison. MAWPS is an easy English dataset where most problems contain only two operands.  Following most previous works~\cite{jie2022learning,zhang2020graph,mwpnas2023}, we combine all the splits and conduct 5-fold cross-validation for MAWPS.
Table~\ref{tab:dataset} shows the details of dataset partition.

\begin{table*}[h]
\centering

\caption{Dataset statistics. MathQA split follows \cite{Tan2021pre} and \cite{Li2022aclcl}; MathQA$\dagger$ follows \cite{Li2022aclcl}.}
\begin{tabular}{c|cccc}
\hline
\textbf{~~Dataset~~}    & \textbf{~Training~} & \textbf{~Validation~} &  \textbf{~Testing~} &  \textbf{~Language~}          \\ \hline
~~MAWPS~~       & 1589   & 199   & 199 & English  \\
~~Math23k~~       & 21162   & 1000   & 1000 & Chinese  \\
~~MathQA~~        & 23703  & 3540  & 2410 & English \\
~~~{MathQA}$^{\dagger}$~~        & 16191  & 2411  & 1605 & English \\ 
\hline

\end{tabular}
\label{tab:dataset}
\end{table*}

Following previous works\cite{jie2022learning,zhang2020graph}, we apply expression accuracy and answer accuracy as the evaluation metrics. Expression accuracy indicates that if the predicted solution expression is the same as annotated expression. Value accuracy indicates whether the final answer calculated from the predicted expression is equal to the gold value. 
Higher scores reveal better performance for both metrics.

\section{Details of Baselines}
As aforementioned, existing works lie in three paradigms: seq-to-seq, tree-decoder and reasoning.
To evaluate the effectiveness of the proposed pseudo-dual learning for MWP solving, we equip an extensive set of baselines with our method, and the details are as follows: 
\begin{itemize}

\item\textbf{DNS} \cite{wang2017deep} develops a vanilla seq-to-seq model to generate expressions, where the encoder is BiGRU and the decoder is LSTM.

\item\textbf{GTS} \cite{xie2019goal} designs a goal-driven tree structure decoder to generate  expression tree by initializing the root node after encoding problem by BiGRU and decomposing goal in a top-down manner.

\item\textbf{Graph2Tree} \cite{zhang2020graph} uses a graph encoder to represent quantities enhanced by constructing quantity cell graph and quantity comparison graph.

\item\textbf{BERT-Tree} \cite{Li2022aclcl} employs a pre-trained language model BERT as word problem encoder and outputs expression by tree decoder.

\item\textbf{RE-Deduction} \cite{jie2022learning} introduces a complex relation extraction framework to deduct expression. Here, we choose Roberta-Base as semantic encoder.
\end{itemize}
\section{Experimental Implementation Details}
\label{sec:setting}
We reproduce all the baselines and implement the PseDual versions depending on the official open-sourced codes. The Adam algorithm \cite{kingma2015adam} is employed  to optimize all the models with initial learning rate $\{$1e-3, 1e-4, 5e-5, 2e-5$\}$ for different models, which are the same as the ones reported in corresponding papers.
For expression encoder of the reexamining module, the token (\myie{}, operands and operators) embedding dimension is set to 128 and the dimension of hidden state is equal to word problem encoder. 
The temperature $\tau$ of gumbel softmax is annealed to 0.5 at a rate of 3e-5 every 100 iterations in each epoch. 
The weight coefficient $\epsilon$ between target expression and generated expression decays exponentially at a rate of 0.99999.
For fair comparison, we set the maximum training epochs  the same as original work and choose the best model verified on validation set, then report the performance on test set.
During testing, beam search  strategy with size $\{$1, 3, 5$\}$ for different baselines (according to corresponding paper) is utilized. 
All the experiments are conducted on a workstation with 8 Titan V.

\end{document}